\documentclass[lettersize,journal]{IEEEtran}

\usepackage{amsmath,amsfonts}
\usepackage{algorithmic}
\usepackage{algorithm}
\usepackage{array}
\usepackage{textcomp}
\usepackage{stfloats}
\usepackage{url}
\usepackage{verbatim}
\usepackage{graphicx}
\usepackage{cite}
\usepackage[english]{babel}
\usepackage[utf8x]{inputenc}
\usepackage{multicol}
\usepackage{breakcites}
\usepackage{comment}
\usepackage[colorlinks=true, allcolors=blue]{hyperref}
\usepackage[capitalize]{cleveref}
\usepackage{booktabs}
\usepackage{enumitem} 
\usepackage{csquotes}
\usepackage{lineno}
\setlist[itemize]{noitemsep}
\usepackage{subcaption}

\usepackage{framed, multido}

\newcommand{\definitionbox}[2][6]{
\begin{framed}
\noindent#2%
\end{framed}}

\usepackage{pgfplots}
\pgfplotsset{compat=1.17}
\usepgfplotslibrary{fillbetween}

\usepackage{tikz}  

\definecolor{prussianblue}{RGB}{4,35,58}
\definecolor{pennred}{RGB}{153,0,0}
\definecolor{lapislazuli}{RGB}{63,100,147}
\definecolor{sage}{RGB}{164,180,148}
\definecolor{earthyellow}{RGB}{255,165,82}

\definecolor{hookersgreen}{RGB}{89,130,122}
\definecolor{pearl}{RGB}{224,220,194}
\definecolor{slategray}{RGB}{154,181,196}
\definecolor{bittersweetshimmer}{RGB}{204,68,75}

\definecolor{coral}{RGB}{252,122,87}
\definecolor{lightcoral}{RGB}{242,132,130}
\definecolor{burntsienna}{RGB}{238,108,77}
\definecolor{skymagenta}{RGB}{179,123,164}

\colorlet{creationcolor}{sage!30}
\colorlet{curationcolor}{earthyellow!30}
\colorlet{schedulingcolor}{slategray!30}
\colorlet{evaluationcolor}{lightcoral!40}

\colorlet{accuracycolor}{prussianblue}
\colorlet{biascolor}{pennred}
\colorlet{diversitycolor}{lapislazuli}
\colorlet{consistencycolor}{hookersgreen}
\colorlet{relevancecolor}{skymagenta}

\colorlet{bandaranzalicolor}{slategray}
\colorlet{mexicocitycolor}{hookersgreen}
\colorlet{blackforestcolor}{prussianblue}
\colorlet{mumbaicolor}{lightcoral}
\colorlet{capetowncolor}{bittersweetshimmer}
\colorlet{chabarovskcolor}{skymagenta}
\colorlet{kipparingcolor}{pearl}

\colorlet{sourcedomaincolor}{prussianblue}
\colorlet{targetdomaincolor}{burntsienna}

\usetikzlibrary{
  arrows,       
  calc,         
  positioning,  
  shapes,       
  backgrounds,  
  decorations,  
  patterns,     
  intersections,
  fit           
}

\pgfdeclarelayer{background}
\pgfdeclarelayer{foreground}
\pgfsetlayers{background,main,foreground}

\pgfdeclarelayer{background}
\pgfdeclarelayer{foreground}
\pgfsetlayers{background,main,foreground}

\begin{document}

\title{Better, Not Just More: Data-Centric Machine Learning for Earth Observation}

\author{Ribana Roscher, Marc Rußwurm, Caroline Gevaert, Michael Kampffmeyer, Jefersson A. dos~Santos, \\Maria Vakalopoulou, Ronny Hänsch, Stine Hansen, Keiller Nogueira, Jonathan Prexl, Devis Tuia
\thanks{}
}

\markboth{}%
{Roscher \MakeLowercase{\textit{et al.}}: }


\maketitle

\begin{abstract}
Recent developments and research in modern machine learning have led to substantial improvements in the geospatial field. Although numerous deep learning architectures and models have been proposed, the majority of them have been solely developed on benchmark datasets that lack strong real-world relevance. Furthermore, the performance of many methods has already saturated on these datasets. We argue that a shift from a model-centric view to a complementary data-centric perspective is necessary for further improvements in accuracy, generalization ability, and real impact on end-user applications. Furthermore, considering the entire machine learning cycle—from problem definition to model deployment with feedback—is crucial for enhancing machine learning models that can be reliable in unforeseen situations.
This work presents a definition as well as a precise categorization and overview of automated data-centric learning approaches for geospatial data. It highlights the complementary role of data-centric learning with respect to model-centric in the larger machine learning deployment cycle. We review papers across the entire geospatial field and categorize them into different groups. A set of representative experiments shows concrete implementation examples. These examples provide concrete steps to act on geospatial data with data-centric machine learning approaches.
\end{abstract}



\section{Introduction}
\label{sec:introduction}

\begin{figure*}[ht]
    \centering
    \includegraphics[width=0.8\textwidth]{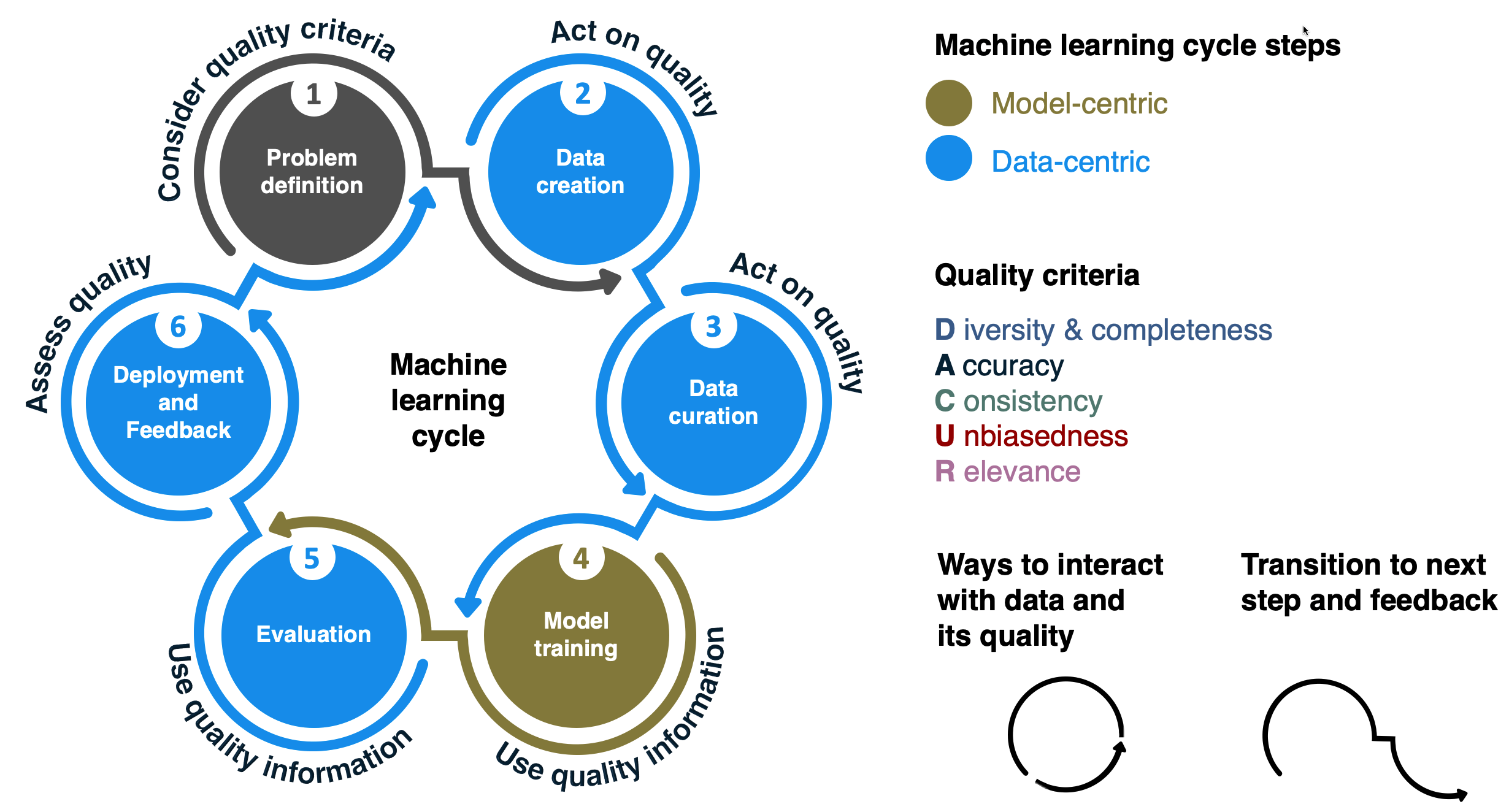}
    \caption{Steps of the machine learning cycle. Each step highlights one way to interact with data and its quality, and each step can employ multiple techniques to perform the interaction (see Section~\ref{sec:steps}). It involves problem definition, data creation and curation, model training and evaluation, and final deployment, which feeds back into a modified problem definition. Model-centric learning focuses primarily on model training and evaluation, while data-centric learning involves algorithms covering data curation, creation, specific training strategies, evaluation, and deployment feedback. The considered quality criteria are diversity and completeness, accuracy, consistency, unbiasedness, and relevance, see Section~\ref{sec:terminology}). 
    } 
    \label{fig:pipeline}
\end{figure*}

\IEEEPARstart{R}{emote} sensing data is a central link between pressing global challenges and the possibilities of machine learning methods. To realize its full potential, a deep understanding of the data and its utilization possibilities is essential.
Generally, data plays a fundamental role in the machine learning (ML) cycle with steps from informing (1) problem definition, (2) data creation, (3) data curation, enabling (4) model training, and (5) evaluation to (6) the eventual model deployment that feeds back to modifying the problem definition. We show this cycle in \cref{fig:pipeline} where each node is colored by its focus of being problem-centric (dark gray), data-centric (blue), or model-centric (dark green). 
Yet, current research in machine learning is predominantly model-centric and focuses on model design and evaluation (Step~4). 
This primarily emphasizes optimizing the accuracy and efficiency of the models themselves \cite{aroyo2022data,Polyzotis2021,wagstaff2012machine} and considers the dataset rather as a static benchmark than a dynamic representation of the application. 
Even in applied machine learning areas such as geospatial data analysis, the research towards integrating new machine learning methods has recently largely focused on refining the algorithms and fine-tuning the model parameters \cite{zhang2022artificial}. Data-centric aspects like data acquisition (Step~1) and  curation (Step~2) are done manually to a large degree and data quality is rarely considered. At the same time, most domain expertise remains concentrated in classic feature engineering.

The rapid development of machine learning methods has been primarily driven by the availability of large-scale datasets and advancements in computing power \cite{kraljevski2023limits,schmitt2023there}. This has allowed researchers to focus on developing complex models capable of capturing patterns and relationships within the data. 
However, many developments have been pursued in controlled benchmark settings only, where the natural variability and real-world characteristics of the data are bypassed. In the geospatial domain, for example, this is achieved through a one-time pre-defined data location sampling, as seen in the predefined regions of the Sen12MS dataset \cite{Schmitt2019}, or through class-balancing techniques commonly used in crop-type mapping \cite{garnot2021panoptic}. Although the great importance of these benchmarks should not be diminished, as a result, many methods developed are often detached from reality and seldom robust in real-world deployment. This requires considering all steps in the machine learning pipeline and acknowledging it as a cycle (\cref{fig:pipeline}), as well as a high degree of robustness for acquisition specification and generalization to diverse shifts.

Despite the availability of enormous amounts of data \cite{schmitt2023there}, including well-curated benchmarks, and significant progress made through methodological advances, there is currently a saturation point where many established architectures and training methods achieve comparable accuracy (\cite{goldblum2023perspectives}, 2023 AI Index Report\footnote{\url{https://aiindex.stanford.edu/report/}}). This is exemplified by foundation models or general-purpose computing architectures for multiple modalities and tasks \cite{hong2024spectralgpt,wang2024hypersigma,mai2023opportunities,sun2022ringmo}. These large-scale models, trained on massive but mainly domain-agnostic datasets, increasingly replace modality-specific recurrent and convolutional models. This development comes with the prevailing belief that "more data is better". However, the reality is more complex. 
While larger datasets can provide a broader representation of real-world scenarios, foundation models require adaptation through downstream models tailored for specific applications through fine-tuning datasets that can be substantially smaller in size. 
Still, the smaller the size, the more important the data quality becomes, as noisy labels and other errors can have increasingly detrimental effects on final accuracy.  
In this direction, it has already been argued that in addition to the quantity of data, the quality is a significant factor in model performance across various domains \cite{ghamisi2024responsible,ng2022unbiggen,jarrahi2023principles,aroyo2022data,sambasivan2021everyone,li2021cleanml}. 
That means that not all data samples contribute equally to model learning and generalization. In fact, indiscriminately increasing data volume can lead to diminishing returns or even detrimental effects on model performance. 
The assumption that more data inherently leads to better outcomes overlooks the complexities of data distribution, the potential for introducing biases, spurious correlations, and the computational resources required for processing and storing vast datasets \cite{falk2023challenging,hadsell2020embracing}. Moreover, many scientific disciplines, particularly those reliant on experimental data, face significant data scarcity. In these domains, the emphasis is on quality rather than quantity, as each data point is meticulously collected and holds substantial scientific value. Therefore, while foundation models represent a leap forward in AI capabilities, the general belief that "more data is better data" does not fully address the complexities of real-world applications. Effective solutions require an approach that considers both the quality and quantity of data, as well as the adaptability of models to specific tasks and domains.

However, a number of challenges need to be overcome before this can be turned into actions.
The notion of data quality and its specific criteria is only vaguely defined yet and can differ depending on the domain and task. In the geospatial domain, we have identified five criteria especially relevant to machine learning: diversity and completeness, accuracy, consistency, unbiasedness, and task-relevance, which will be discussed in this paper.
Automating to interact with the data regarding their quality in each step in the machine learning cycle is crucial. This is especially important when dealing with training and evaluation datasets that are too large for traditional visual refinement and manual data manipulation.
The importance of shifting to a more data-centric perspective is also underlined by the growing demand for methods that are robust to noise \cite{aksoy2022multi,fatras2021wasserstein} or missing data ~\cite{zhang2018multi,chen2021practical,machado2023missdata}, or that are specifically designed to handle data quality deficiencies. Here, it becomes clear that assessing and optimizing data quality is crucial in addressing real-world data challenges.
However, such methods represent only one way to interact with data regarding their quality (Step~4: use quality information during model training).

This article offers a systematic discussion of data-relevant aspects of the machine learning cycle in the context of geospatial data. 
It has the goal of fostering awareness about the benefits of optimizing geospatial data alongside new methods of research. 
\noindent
Within the scope of this paper, we use the term
\definitionbox[1]{\textbf{data-centric learning} as a paradigm focusing on the systematic, automated, and algorithmic determination, as well as the utilization of a rich and high-quality dataset, including a rigorous evaluation process to ensure that the model performs optimally on the dataset for the intended task.}
In contrast, model-centric learning focuses on the optimization of the model parameters that include trainable parameters and hyperparameters, such as the model design, as well as the learning objective as a loss function. It also focuses on the definition of evaluation metrics to measure the model's performance with respect to a static evaluation dataset.

The article is organized as follows: It first clarifies the terminology, definitions, and goals of data-centric and model-centric learning (\cref{sec:terminology}) and then provides a comprehensive discussion of data-centric machine learning and its steps, with a focus on the analysis and interpretation of geospatial data.
Additionally, it presents a systematic review of data-centric machine learning techniques tailored explicitly to geospatial data (\cref{sec:steps}), facilitating the learning and evaluation of high-performing machine learning models (see \cref{fig:pipeline}). 
The paper is supported by three experiments using data-centric machine learning techniques, presented in \cref{sec:experiments}.

\section{Definitions and Terminology}%
\label{sec:terminology}

\noindent
This section describes the machine learning cycle, provides definitions of data-centric and model-centric learning, and details the pursued data quality criteria. We refer to data as a set of individual raw samples and their targets, where a data sample may be assigned to one or more targets.

\subsection{Machine learning cycle}
The machine learning cycle, illustrated in \cref{fig:pipeline}, involves the following sequential steps that may be iterated following the entire circle or within intermediate steps:
\begin{enumerate}

\item \textbf{Problem definition.} Definition of the underlying objectives by leveraging problem-centric expertise through domain knowledge. Methodologically, this step defines the ML task, e.g., classification or regression, the method, data type and sources based on the application's goals, needs, and restrictions. 

\item \textbf{Data creation.} 
Generation or collection of data from various sources. This includes manually collecting data through surveys or measuring instruments, automatically acquiring sensor data, systematically and algorithmically scraping data from the Web, or generating data using generative models. 

\item \textbf{Data curation.} Utilizing curation strategies to obtain high-quality data and prepare data for modeling. This can be done through outlier filtering, label noise identification and reduction, or de-biasing data with respect to relevance to, for instance, the region of interest.

\item \textbf{Model training and hyperparameter tuning.} Learn and fine-tune a model utilizing the curated data for training and validation. This involves choosing the model architecture and objective function.

\item \textbf{Evaluation.} Applying the machine learning model to evaluation data to assess model performance. This involves, in close connection to Steps~1 and~2, the creation and curation of evaluation data or defining a suitable test split that may be spatially distant from training areas. 

\item \textbf{Model deployment, monitoring, and feedback.} Continuously monitoring the trained model regarding its performance in previously unknown real-world applications and giving feedback to improve the data and the model. 
This step generates feedback from the model deployment and the monitoring to inform about changed needs and goals and to improve the overall process. 
\end{enumerate}

\noindent Within the cycle, data-centric learning focuses on Steps~2: \textit{Curation}, 3:~\textit{Creation}, 5:~\textit{Evaluation}, and 6:~\textit{Deployment}.
In contrast, model-centric learning involves Steps~4: \textit{Model training} and~5: \textit{Evaluation} in terms of metric definitions.

\subsection{High-quality criteria for geospatial data}
While model-centric learning approaches focus on the optimization of performance on a static evaluation set, data-centric approaches focus on a series of quality criteria of the data used. We identified five criteria that are defined in the following. We further indicate whether each criterion is typically considered at a global (dataset-wide) level or a local (individual data sample or subset) level. Please note that some criteria have already been pointed out or other criteria might be more important in specific applications (e.g., \cite{long2021creating}). Moreover, criteria can be closely connected or partly overlap in specific settings, however, we consider them as distinct concepts that describe different characteristics of the data. This is demonstrated, for example, by the fact that if one quality criterion is high, other criteria do not automatically have to be high as well. We consider the data to be observations and labels and do not make an explicit distinction here.

\begin{itemize}
    \item \textbf{Diversity and completeness (global).} Completeness refers to the extent to which all necessary data samples are present within the dataset. It involves ensuring that there are no missing data points and that the dataset includes all relevant information required for the specific task. In the geospatial domain, completeness is critical as missing data can arise due to sparse data acquisition, multimodal fusion with different resolutions, or challenges in gathering information from different sensors. Techniques to measure completeness include data coverage analysis and gap analysis, while metrics can include the percentage of missing values and the spatial coverage ratio. 
    Closely related, diversity pertains to the inclusion of a wide range of data samples that represent various scenarios and variations, enabling ML models to handle various real-world scenarios \cite{shankar2017classification}. In the context of geospatial data, this means including data from various geographic areas, different times, and multiple sensor types to capture the full range of potential conditions. It is important to consider diversity within the context of completeness, which ensures that all necessary information is covered without important details missing. 
    
    \item \textbf{Accuracy (local).} Data accuracy indicates how closely the data values align with the true or expected values. It quantifies the correctness and precision of the data, accounting for any errors or deviations that might occur during data collection. In the geospatial domain, accuracy is influenced by factors such as random variations or errors during data acquisition, sensor precision, environmental conditions, and data processing techniques \cite{gawlikowski2023survey}. Additionally, human errors in generating reference labels \cite{elmes2020accounting}, or situations where the phenomenon targeted by machine learning models is conceptually ambiguous, such as identifying informal settlements from satellite imagery \cite{gevaert2019challenges} or classifying wilderness \cite{ekim2023mapinwild,stomberg2023recognizing}, contribute to uncertainty. Accuracy can be, for example, assessed by repeated measures or comparing the data against reference values.

    \item \textbf{Consistency (local and global).} Consistency, locally, refers to the uniformity of properties or attributes of single data samples within one dataset and across multiple datasets. Globally, it ensures that the dataset maintains uniformity in terms of data formatting, units, and definitions across all samples. For geospatial data, consistency is crucial when integrating data from different sources, such as varying coordinate systems or temporal resolutions. In the case of machine learning tasks for geographic data, ensuring consistency is critical during the process of annotating samples for training models \cite{fenza2021data,vargas2020openstreetmap}. 
     
    \item \textbf{Unbiasedness (global)}. Unbiasedness measures the extent to which data is free from systematic errors and and consistent deviations. An unbiased dataset accurately reflects the true values and distributions without any systematic over- or underestimation. In the geospatial domain, biases can be manifold \cite{ghamisi2024responsible}. For example, acquiring data is generally easier in data-rich geographic areas such as Europe or North America. However, it can be much more challenging in other continents with limited data availability, which can introduce bias \cite{ghamisi2024responsible,shankar2017classification}. 
    
    \item \textbf{Relevance (local).} Relevance denotes how applicable and suitable a data sample is for the specific task or problem being addressed. It assesses the alignment of the data with the requirements and objectives of the analysis or model. For geospatial data, relevance involves ensuring the data is up-to-date and reflect the present conditions of the Earth. As already pointed out by~\cite{Polyzotis2021}, datasets and models need to be updated continuously to be relevant.  
    
\end{itemize}

Generally, having measures for the data quality allows quantifying the goodness of a single sample (locally) or the whole dataset (globally) with respect to a given criterion, which provides the user with intuition and builds the basis for actions in data-centric approaches. A measure also helps identify issues in the dataset that would hinder the development of a well-performing model. Once the quality is quantified, specific steps can be taken to address the identified issues and modify the dataset. Quality measures serve as a guide for creating, collecting, enhancing, and curating the data. Additionally, a quantitative measure enables us to prioritize the steps involved. Existing approaches to quantify the quality (value) of data can be divided into label-based~\cite{ghorbani2019data,wang2023data} and label-free approaches~\cite{xu2021validation}. Label-based approaches typically evaluate the effect that a given sample has on the validation loss. In contrast, the label-free approaches leverage inherent data characteristics to assign values according to the desired quality criteria.
While a quantification of quality can find direct applications in the form of data marketing~\cite{schomm2013marketplaces}, it more often is used to facilitate downstream tasks such as active learning ~\cite{tuia2011survey} or data cleaning~\cite{ghorbani2019data}, where the value is used to select relevant or prune irrelevant data samples.

\section{Data-centric machine learning techniques}%
\label{sec:steps}

\begin{figure*}[p]
    \centering
    \resizebox{\textwidth}{!}{
    \input{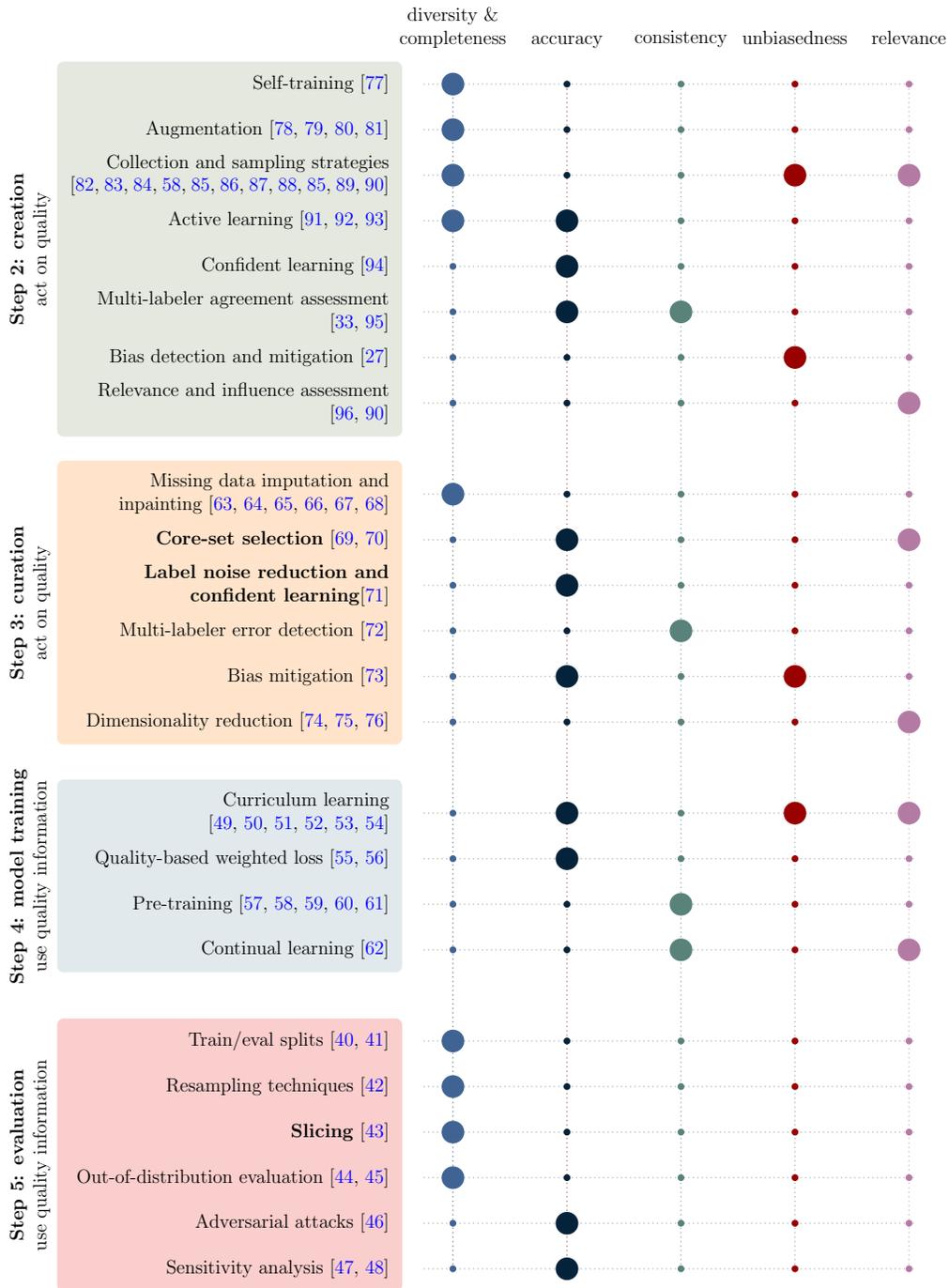}
    }
    \caption{Data-centric machine learning techniques and papers referenced in \cref{{sec:steps}}. Each machine learning step and each technique (rows) interacts in a specific way with the data and the quality (columns). The techniques used for our experiments are highlighted in bold. Large dots indicate which quality information is used (model training and evaluation), or which specific quality criteria is acted on (in the creation and curation step).}
    \label{fig:techniques_dotplot}
\end{figure*}

In this section, we thoroughly review the literature and present data-centric techniques for the machine learning Steps~2-5 with a specific focus on geospatial remote sensing data. We skip Step~1, since it mainly contains manually performed actions, and we skip Step~6 since it mainly is concerned with the feedback loop. We group and structure these techniques in \cref{fig:techniques_dotplot} with respect to the mentioned quality criteria. Each step in the machine learning cycle interacts with data in a specific way regarding its quality. This includes acting on data quality, using quality information, assessing quality, and considering the quality. 
The review aims at consolidating knowledge helping to quickly understand the current state of the field, building the basis for dentifying gaps and opportunities for future research.

\subsection{Data Creation (Step 2)}
Data creation, whether through collection or generation, directly acts on the data quality. It serves two distinct purposes: firstly, high-quality training data supports the model learning process, while secondly, high-quality evaluation data enhances the information content to provide deeper insights into the model's capabilities (see also Step~5: \textit{Model evaluation}). 

\textbf{Diversity and completeness.} 
Fostering diversity and completeness are closely related to the sampling scheme, which should ensure heterogeneity, such as across different geographic regions (spatial diversity) or related environmental conditions, for example, to seasonal variations (temporal diversity). Measuring the heterogeneity of samples in geospatial applications can be based, for example, on spectral similarity \cite{Geiss2022,Castillo-Navarro2022} or morphological image characteristics \cite{Gevaert2022}. Wesley et al.~\cite{Wesley2021} propose the Geospatial Inception Score, which considers diversity as a function of the co-occurrence of samples, their variety, and their context. Another measure is presented by Xu et al.~\cite{xu2021validation}, who propose a quantification of diversity in terms of the volume of the data matrix (determinant of its left Gram matrix). A sample's contribution to diversity can then be measured via the increase in volume when adding it to the existing dataset.
Generally, sampling schemes should be designed to cover the expected diversity of the classes and to avoid biases. For example, stratified random sampling is a long-standing good practice in geospatial applications \cite{OLOFSSON201442,Stehman2009}. 
Further efforts to improve diversity and completeness involve techniques to merge data from multiple sensors (e.g., optical, radar, LiDAR, or thermal) and collecting data at different spatial scales to create more comprehensive datasets \cite{li2022deep,zhu2022novel,li2022deep,persello2014active}. This ensures that data encompasses a wide range of features, supporting an understanding of the geographical area under study. However, research in this direction has so far been the exception rather than the rule and a close interaction with model-centric developments is necessary to find suitable strategies to integrate multi-resolution, multi-temporal, and multi-spectral data in a joint framework \cite{ghamisi2024responsible,rolf2024mission}.  

Moreover, it is not always feasible to collect a sufficient amount of data. This is particularly the case if the reference data is collected via in-situ measurements which is often time-consuming and a high effort \cite{fowler2020all}. To address this limitation, data augmentation has become a prominent data-centric technique. This technique involves artificially increasing the size of the dataset by applying various geometric and spectral transformations to the input data samples \cite{wang2018sar}. Also, generative adversarial networks (GANs) and diffusion models are increasingly used to generate synthetic data samples and augment the existing data. This helps to diversify the dataset by creating samples coherent with the original data distribution \cite{mumuni2022data,mansourifar2022gan,zhang2021stagewise}. Although the general concepts are promising, research gaps comprise the kind of augmentation to ensure a physically consistent model and the capturing of powerful features from the data \cite{burgert2024estimating,willbo2024impacts}. 

If a learned model is already available, it can be used to acquire new data through self-training \cite{TONG2020111322}. In self-training, the trained model is used to classify unlabeled samples, and the predictions are used as pseudo-labels. Samples with their pseudo-labels are then converted into labeled training data based on specific quality criteria. Another technique is active learning \cite{tuia2009active}. Here, the focus is on identifying unlabeled data samples that would be most useful and should be added to an existing dataset. The labeling is performed manually, a process known as human-in-the-loop. Used metrics in this context are the Most Ambiguous and Orthogonal (MAO) metric \cite{ferecatu2007interactive} and angle-based diversity \cite{demir2010batch} that utilizes the angle between two samples in a feature space to measure diversity. However, the selection of samples is still an open research question \cite{kirsch2019batchbald} and current research point towards a necessity to use multiple quality criteria beyond diversity and completeness, so far mainly focusing on uncertainty.

\textbf{Accuracy.} 
Fang et al. \cite{fang2022confident} employ confident learning \cite{angluin1988learning} to select high-confident training samples in a domain adaptation setting from a set of pseudo-labeled samples in the target domain scene. The confidence of the labels is assessed by comparing their reference label with the estimated score for that label. A remaining challenge, however, is the notion of data accuracy and confidence and their disentanglement from model uncertainty, with uncertainty quantification being a research area gaining increased attention \cite{gawlikowski2023survey}. Bastani et al. \cite{bastani2023satlaspretrain} conduct iterative analyses during the design of their {SATLASPRETRAIN} dataset to evaluate the precision and recall of labels collected from various sources (e.g., manually labeled and from existing products). This process helps improve the overall quality of the labels. Additionally, they provide information on the accuracy of the categories determined during dataset creation.
With knowledge about the problem-specific patterns, label refinement strategies can be employed, for example, by improving coarse hand-annotations. Here, for instance, Ru\ss{}wurm et al. \cite{russwurm2023large} designed a simple computer vision pipeline to refine coarse hand-labels of floating plastic marine debris from Sentinel-2 images. This label refinement improved the overall model accuracy of the segmentation model.

\textbf{Consistency.} 
Data consistency is maintained by implementing rigorous data collection procedures, which include standardizing measurements or labeling procedures, and quality control measures. Ensuring data consistency can be particularly challenging when multiple experts label reference data, such as through crowdsourcing platforms~\cite{zhang2020crowd}. While online crowdsourcing platforms like OpenStreetMap (OSM) facilitate the labeling of many samples, the data quality is not always guaranteed, which leads to consistency issues \cite{burgert2022effects}. To address this, the level of agreement among labelers is often used as a measure of consistency. For example, Fenza et al. \cite{fenza2021data} mention that simple measures, like the ratio of agreeing labelers to all labelers, can be used as an indicator of consistency. This measure can be further adjusted by considering the uncertainty and reliability of the labelers \cite{gomes2011crowdclustering}. Future directions and research gaps have been identified, for example, by Saralioglu and Gungor \cite{saralioglu2020crowdsourcing}.

\textbf{Unbiasedness.} 
Data creation imposes three types of bias: historical bias, representation bias, and measurement bias ~\cite{ghamisi2024responsible, suresh2021framework}. 
Influencing factors for the choice of specific sampling strategies may include logistical and practical reasons for in-situ data collection. For example, as mentioned by Fowler et al. \cite{fowler2020all}, in-situ crop-type labels are often collected in close proximity to roads. Even though diversity and completeness can be of high quality, biases can occur. Self-supervised learning methods like seasonal contrast \cite{manas2021seasonal}, for example, take special care in sampling (unlabelled) images geographically according to some distribution (e.g., population density) to cover all possible representations of the world. A naive uniform sampling, though ensuring diversity to a high extent, would generate too many examples over repetitive uninformative areas like deserts or oceans. It is important to note that class imbalance in training data is a common challenge in geospatial data, which can cause the model to have a bias towards the majority class in classification tasks. This imbalance may naturally occur in some geospatial applications, where the distribution of objects or features in the studied geographic area leads to imbalanced classes. For example, in land cover classification, urban areas may be a minority class compared to forests or water bodies. 
Finally, measurement bias is related to how the labeling scheme is conceptualized. For example, because of the ambiguity of the class definitions, rules for labeling will always introduce bias in how objects are identified. This cannot be avoided but should be clearly communicated. Assessing and mitigating biases is a so far underrepresented research area, which however gained attention with the movement from local to global models \cite{ghamisi2024responsible}.

\textbf{Relevance.} 
Without an existing model, there are limited works that consider a relevance measure for data creation. Instead, other quality criteria are primarily used. As a result, the collection of relevant data heavily relies on the decisions made in the problem definition step (Step~1). An approach to improve data relevance is tested in Ru\ss{}wurm et al. \cite{russwurm2023detection}, who aimed to improve urban settlement detection in Mozambique and Tanzania by dynamically sampling training data of urban areas in proximity to the study region. 

To quantify the relevance of data to a given task when a model is available, one direct approach is the "leave-one-out" method~\cite{cook1977detection}, where the model's performance, trained on the entire dataset, is contrasted with the model's performance when a particular data point is excluded. The difference in performance can then be interpreted as the data point's relevance for the task. However, computing such values for all data points using this method can be excessively resource-intensive. A more efficient alternative to address this challenge is the use of influence functions~\cite{koh2017understanding,chen2021hydra}, which assess the influence of a specific sample on the model's parameters when that sample's weight is increased. To further address the complexity of sample interactions, the concept of Data Shapley has emerged~\cite{ghorbani2019data}, drawing inspiration from Shapley values. Data Shapley provides a more comprehensive framework for evaluating the value of individual data points in a dataset and efficient data Shapley approaches have been proposed~\cite{ghorbani2020distributional}. However, such methods have not been used for geospatial data so far. In the context of incremental learning in the geospatial domain, Roscher et al. \cite{roscher2012i2vm} uses the classification model to identify relevant samples based on their ability to help the model to adapt to the new data. 
More techniques that rely on an existing model are discussed in \cref{sec:model}.

\subsection{Data Curation (Step 3)}
Data curation refers to the systematic and algorithmic refinement of a created dataset. We specifically focus on techniques that remove or correct data. The main objective is to improve data quality, which in turn should benefit model efficiency, accuracy, and generalizability. Similar to the data creation step, the data curation step directly impacts the quality of the data according to our defined criteria.

\textbf{Diversity and completeness.} 
A commonly used set of techniques is the filling of gaps in spatial and temporal coverage, as missing data is a common problem and causes different issues such as performance degradation, problems in data analysis, and biased outcomes~\cite{emmanuel2021survey}. Besides classical interpolation strategies, a recent approach is the use of generative models like generative adversarial networks. Luo et al. \cite{luo2018multivariate}, for example, impute data in one-dimensional time-series signals and outperform classical methods, and Dong et al. \cite{dong2018inpainting} perform inpainting in sea surface temperature images where clouds cover the scene. 
Another set of techniques that impute missing data is superresolution \cite{wang2022comprehensive}. 

\textbf{Accuracy.} One of the most commonly used sets of techniques is data-cleaning, which specifically focuses on data accuracy and consistency \cite{li2021cleanml}. 
Data-only unsupervised techniques are applied before model training and are independent of the learning objective and application. However, as discussed by Ilyas et al. and Neutatz et al. \cite{ilyas2022machine,neutatz2021cleaning}, methods that clean the data with respect to the application goals and take the ML model into account show more promising results.
Core-set selection techniques such as the one presented by Santos et al. \cite{santos2021quality} can be considered a data-cleaning approach as the selection process retains the cleanest samples. In their approach, they use clustering for satellite time series to identify samples that are mislabeled or have low accuracy with the goal of removing them from the training set to avoid a decrease in model performance. 
In some cases, the source of uncertainty is known and can be directly reduced. For instance, data uncertainty may arise from the presence of clouds, and many methods have been developed to automatically remove them \cite{zhang2019coarse,li2020thin,ebel2022sen12ms} (see also techniques that act on diversity and completeness). 
Northcutt et al. \cite{northcutt2019confidentlearning}, for example, use confident learning \cite{angluin1988learning} to identify and correct label errors. Labels are flagged as noisy when their label uncertainty is high, which means the model is not confident about the label (see \cref{sec:exp2}). These labeled samples are then iteratively re-labeled using a more accurate labeling function, which can be a human annotator or a more sophisticated model.
When multiple labelers provide labels, similar strategies can be used to identify the most likely labels or remove highly confusing samples \cite{long2021creating}.
Most model-based approaches, however, use data with their quality information during model training instead of cleaning the data and re-training the model in two separate steps (see Model training and data utilization (Step 4)).

\textbf{Consistency.} Curation activities further aim to establish consistency across different datasets or data sources and annotators. 
Hechinger et al. \cite{hechinger2023categorising} demonstrate that when a fixed set of labelers is given, there are often 'voting patterns' where labelers agree or disagree in a similar manner. By conducting a clustering analysis of the LCZ42 dataset \cite{zhu2020so2sat}, which was labeled by multiple experts, they gained valuable insights regarding the distinguishability of classes, voting behavior, and the level of consensus based on the geographic location of the reference images. These insights can be incorporated into model training (Step~4) or used to develop curation strategies, such as removing labels with low consensus values. An open challenge is the usage of efficient labeling strategies or the identification of a subset of samples that should be annotated by multiple labelers to keep the effort low. 

\textbf{Unbiasedness.} 
Furthermore, curation aims to detect and eliminate systematic biases or distortions in geospatial data. In cases where biases result from class imbalances or a non-representative spatial distribution \cite{shankar2017classification}, similar to resampling strategies that generate or collect new data, undersampling strategies for the majority class can be utilized. Risser et al. \cite{risser2022survey} discuss methods to remove biases in satellite images, for example, by targeted removal of specific features or samples. 
It is important to note that removing noisy data or data with missing values, as discussed in \cite{emmanuel2021survey}, can introduce further bias, therefore auditing strategies should be used \cite{ghamisi2024responsible} for checking. Moreover, whether class imbalance is desirable or not depends on the goals and requirements of the analysis \cite{johnson2019survey}. 

\textbf{Relevance.} 
Curation also considers the relevance of the data. By removing irrelevant data, curation ensures that the data is tailored to specific needs and goals, focusing on the most relevant information. It also naturally reduces the dataset size, improving data efficiency. 
Core-set selection methods compute the distance between training and test samples and remove those with low proximity. For example, in geospatial air quality estimation, Stadtler et al. \cite{stadtler2022explainable} demonstrate that removing the 10\% of training samples with the lowest proximity to the test samples only slightly decreases test accuracy, as these samples are not relevant for training. However, they assume that the test set is representative enough for the complete data set. A more sophisticated approach is implemented and described in our first experiments (\cref{sec:exp1}). 

In remote sensing applications with high-dimensional data, there is often a focus on dimensionality reduction techniques. These techniques aim to reduce complexity while preserving essential information. Established feature selection techniques reduce redundancy and noise, thereby increasing model performance by focusing on the most relevant and informative features \cite{chen2022comparison,georganos2018less}. With the field of explainable machine learning, new methods are introduced to calculate importance scores, sensitivities, or contributions of features and interpret them as relevance \cite{roscher2020explain}. Combining these areas, Zhang et al. \cite{zhang2023marine} use a wrapper feature selection method and Shapley values to identify the most relevant multi-source remote sensing marine data features used for estimating global phytoplankton group compositions.

\subsection{Model training and data utilization (Step 4)}
\label{sec:model}
Data set utilization is centered around optimizing the way data is used during training to improve model effectiveness. Information about the quality of the data is utilized to enhance model training. Various techniques are employed to enhance training by, e.g.,  speeding up convergence or improve the quality of obtained local optima. In this step, the close interplay between model-centric and data-centric learning becomes apparent, demanding for a joint development of methods for both paradignms.   


\textbf{Accuracy.} 
Curriculum Learning, which performs learning from the easiest to most challenging samples, and its variants~\cite{wang2021survey} have been successfully employed for several tasks in the geospatial domain.
Mousavi et al.~\cite{mousavi2022deep}, for instance, introduce a novel deep curriculum learning method for the classification of PolSAR images by using an entropy-alpha target decomposition strategy to estimate the degree of complexity of each image sample. This measure of complexity is related to data accuracy since the speckle generated during the acquisition of images on vegetation targets creates a variation in the observations. 
Ran et al.~\cite{ran2022unsupervised} propose a curriculum learning-based strategy for unsupervised domain adaptation for a semantic segmentation task by gradually introducing more complex pseudo-labeled samples from the target domain to the training process. The complexity is given by the confidence of the samples, defined by the predicted probability.
Xi~{et al.}~\cite{xi2023multilevel} use a combination of curriculum learning and a weighting scheme for samples in a semantic segmentation task where a domain shift arises due to distinct geographic locations of the source and target domain. Instead of only iteratively selecting the most confident pseudo-labeled samples and adding them to the training data set to adapt to the target domain, they introduce a weighting scheme that balances easy-to-transfer samples with a similar appearance as the target domain and hard classes and samples with low confidence.
Dong et al. \cite{dong2021high} propose a weighted loss function term for a landcover classification task that computes the loss with updated labels. The labels are corrected based on the uncertainty estimate for that label, which is calculated as the entropy with the largest and second-largest probability estimate of the network. 
Similarly, Burgert et al.\cite{burgert2022effects} use self-adaptive training for multi-label remote sensing classification by adapting the targets in the loss with a moving average of the given targets and the class probability estimates for them. 
As for the data creating step, the quantification of uncertainty and the notion of confidence is an open challenge.

\textbf{Consistency}. Some self-supervised learning methods utilize the information about the internal consistency within remote sensing data. Notably, Manas et al.~\cite{manas2021seasonal} define a contrastive loss that matches samples of the same location at different times, Wang et al. use simple augmentation methods \cite{wang2022ssl4eo}. Similarly across modalities, Scheibenreif at al. \cite{scheibenreif2022self} and Prexl and Schmitt \cite{prexl2023multi} use the consistencies between optical and radar data to train a deep learning model. Also, Ayush et al. \cite{Ayush_2021_ICCV} show that the consistency between geolocation and image can be used to pre-train deep learning models.

\textbf{Unbiasedness.} 
Kellenberger et al. \cite{kellenberger2022training} employ curriculum learning to improve training strategies in a habitat suitability mapping task, where class imbalance and bias are the major challenges. In their approach, they counteract the class-imbalance by exposing the model with increasingly difficult samples, which means more imbalanced training scenarios during training. 

\textbf{Relevance.} Yuan et al. \cite{yuan2022easy} employ a self-paced curriculum learning (SPCL)-based model in combination with a weighting strategy for visual question answering (VQA) on remote sensing data. The weights are learned with an importance sampling strategy, where a large importance value means that the sample has a high relevance for the model training.
Another strategy that updates the learned model with a schedule is online/incremental/continual learning. Here, the relevance and therefore the schedule is determined based on the timeliness of the samples with the goal of adapting the model to new samples that became available. In addition to timeliness, geographic relevance can also play a role. Thus, observations are not only influenced by the current imaging conditions but also by the local appearance of land cover or objects. Roscher et al. \cite{roscher2011incremental}, for example, performs a multi-class classification of large neighboring Landsat scenes by incrementally updating the model with observations from neighboring scenes that are labeled through a self-training strategy. Bhat et al. \cite{bhat2023efficient} combine continual learning and curriculum learning by using the new data based on their similarity to the old classes, and additionally focussing on the accuracy of the samples from the old dataset to reduce the effect of noise and to avoid the catastrophic forgetting effect. 

\subsection{Model Evaluation (Step 5)}
At this stage, the model is evaluated regarding efficiency, accuracy, and generalization ability. In general, depending on the application, a variety of metrics can be used for the model evaluation and their interpretation, including different data-specific characteristics such as class imbalance or varying prevalence, which affect the selection of the correct model \cite{varoquaux2023evaluating}. We will focus on methods and tools that are used for the model evaluation with a specific focus on methods that make use of the composition of the data and data quality information.

\textbf{Diversity and completeness.}
Following the creation of a dataset, the customary practice involves dividing samples into training, validation, and testing datasets. This division is crucial for monitoring the model's generalization and addressing common issues like overfitting or underfitting, particularly prevalent in deep learning models. While random or stratified sampling is standard in machine learning tasks, applying this directly to geospatial data can lead to spatial auto-correlation issues among nearby data points in the training and test datasets, potentially overestimating model accuracy \cite{meyer2019importance,karasiak2022spatial}. For instance, the winner of the EMCL/PKDD conference TiSeLaC land cover classification contest utilized only geographical coordinates, omitting individual spectral features in predicting land cover \cite{inglada2018follow-up}. Thus, in geospatial applications, data sampling strategies should prioritize spatial independence. Techniques such as spatial leave-one-out cross-validation or using spatially disjoint scenes for evaluation are essential to mitigate spatial correlations and ensure robust model evaluation \cite{hansch2021trap,karasiak2022spatial}. However, Wadoux et al. \cite{wadoux2021spatial} indicate that proper probability sampling without explicit methods to account for spatial autocorrelation is sufficient.
Lyons et al. \cite{lyons2018comparison} compare different resampling methods for classification and show that such techniques provide robust accuracy and area estimates with confidence intervals.
Techniques to identify representative sample sizes~\cite{foody2009sample} for the different sets should ensure the proper evaluation of the models. In~\cite{blatchford2021determining}, the authors proposed a method for estimating the sample size required to represent the modeled or estimated dataset accurately.

A well-generalizing model should predict with a similar accuracy across the entire evaluation set. However, the model's performance often varies depending on the data distribution, which is not captured by aggregated performance measures. Simply splitting the available dataset into training, validation, and testing sets is not sufficient to ensure a reliable evaluation of the model, especially when the optimization of machine learning methods is performed on a subset of data that may not represent the real data distribution, as mentioned in the previous sections.
To address this issue, slicing algorithms can be used, especially for small datasets, to ensure a valuable partitioning of the data. These algorithms aim to split the data and evaluate the model on relevant sub-populations that share common characteristics. Defining these sub-populations can be challenging, but it can be based on expert domain knowledge, such as using slices with similar characteristics like geographic regions~\cite{yokoya2018open}. 

Furthermore, evaluating in and out-of-distribution is also an essential aspect of evaluating the generalization ability of the models. In-distribution evaluation is the most common way to assess the performance of the models since the predefined splits usually correspond to the same source of data, even if this set could include multiple sources and acquisition parameters. Several works discuss the usefulness and applicability of evaluation metrics \cite{pontius2011death,foody2020explaining,rolf2023evaluation}.
Evaluating and challenging the performance of the models on out-of-distribution samples gain more and more attention from the community with publicly available datasets such as the WILDS~\cite{koh2021wilds}, including also Earth observation datasets, being introduced to benchmark the performance of the algorithms on different distribution shifts. In a similar direction, the 2017 IEEE GRSS Data Fusion Contest~\cite{yokoya2018open} aimed at addressing the problem of local climate zone classification based on multitemporal and multimodal datasets and checking the performance of the algorithms on new and separate geographical locations. Meyer et al.~\cite{meyer2021predicting} propose the area of applicability of a trained model by measuring and thresholding the similarity between the training data and potentially test data samples.
Overall, identifying and understanding distribution shifts is a very active research area.

\textbf{Accuracy.} 
Leveraging accuracy-related information of data can be used to assess the robustness and sensitivity of machine learning models, for example, through the adoption of test-time data augmentation \cite{gawlikowski2023survey} or perturbations in the data. Such techniques introduce variations and uncertainties in the data during the evaluation phase offering an evaluation of the model's performance beyond accuracy. Li et al. \cite{li2020error} introduced synthetic label noise to assess the model's robustness to noise inputs. Mei et al. \cite{mei2023comprehensive} provides a review of works that apply different adversarial perturbations to the remote sensing input images to assess the model's robustness against adversarial attacks. Kierdorf et al. \cite{Kierdorf2023} cluster saliency maps obtained from systematically perturbed images and the impact of these perturbations on the model decision to get insight into the reliability of the model.

\section{Validation studies}%
\label{sec:experiments}

The validation studies use three data-centric machine learning algorithms on a common land cover classification problem from satellite imagery: relevance-based sample weighting to compensate for geographic domain shift, label noise reduction and confident learning, and slicing for a comprehensive evaluation.
Each of them highlight a different aspect of data-centric learning and provides both qualitative images highlighting the effect of the procedure and quantitative results that the respective algorithm has on the land cover classification problem. 
They confirm the functionality of existing approaches within the surveyed field and should emphasize that even single data-centric actions in the machine learning pipeline can lead to better models.

\subsection{DFC2020 Land Cover Classification and Experimental Setup}
\label{sec:expsetup}

We chose land cover classification to illustrate different data-centric approaches and use the IEEE GRSS Data Fusion Contest 2020 (DFC2020) dataset \cite{dfc20outcome}. The DFC2020 addressed large-scale land cover mapping with weak supervision. The training data is Sen12MS \cite{Schmitt2019}, which contains only very low-resolution MODIS-derived annotations. However, for the additional validation data (i.e., not contained in the Sen12MS dataset), semi-manually generated high-resolution labels are provided. We use the Sentinel-2 images from this part of the DFC2020 dataset for the following experiments. The Sentinel-2 images are provided with 10 bands (where the native resolution of 10m and 20m is resampled to 10m) and are acquired over seven globally distributed regions. They were semi-manually annotated at high-resolution (10m) according to a simplified International Geosphere-Biosphere Programme (IGBP) classification scheme with eight classes "Forest", "Shrubland", "Grassland", "Wetlands", "Croplands", "Urban/Built-up", "Barren", and "Water".
While DFC2020 is originally a segmentation dataset, we simplify it for classification by taking the most common labelled pixel as the single label for each image.

This dataset offers interesting characteristics that enable us to showcase different aspects of the potential of data-centric learning. On the one hand, the image data is spatially well distributed over different climate zones, biomes, and socio-cultural regions. This results in data characteristics that will be shared only among a subset of images. A similar diversity can also be expected from the semantic content of the data as captured by the annotations. On the other hand, parts of the data will be easier to correctly classify than others as they are better represented by prototypical samples and are more homogeneous in their appearance (e.g. samples of the "Barren" or "Water" class). Furthermore, while the annotations are created in a semi-automatic process, i.e. with thorough quality control, they still contain a certain amount of noise, ambiguity, and errors.

The seven scenes are split into non-overlapping $256 \times 256$ pixel patches, resulting in a total of 6,114 images. As we are dealing with classification tasks for the following experiments, we follow the classification-oriented conversion proposed by the creator of dataset and determine the dominant label (i.e., the class with the most pixels) within a patch and use it as its patch-level label \cite{Schmitt2021}. These samples are partitioned randomly into 4,000 images for training, 1,128 for validation, and 986 images for testing in the Experiments~2 and 3. In Experiment~1, we split the images by region to specifically mitigate and correct for the distribution shift between regions.
In this first experiment, we also use a common Random Forest classification model, while for Experiments~2-3, we use a deep classification baseline model and train a ResNet-18 convolutional neural network with the AdamW optimizer using a learning rate of 0.001 and weight decay of $10^{-5}$ with up to 1000 epochs. We saved the model with the lowest validation loss, which resulted in a model with 123 epochs after four hours of training.

\subsection{Study 1: Relevance-based sample weighting to compensate for geographic domain shift}
\label{sec:exp1}
\begin{figure*}[t]
    \centering
    
    \begin{subfigure}{.49\textwidth}
        \newcommand{\legend}{%
\begin{tikzpicture}[xscale=1.8, yscale=0.25]
\node[circle, fill=bandaranzalicolor, label={[font=\scriptsize\sffamily, xshift=0.15em]right:Band. Anz.}, inner sep=0.2em] at (0,0){};
\node[circle, fill=kipparingcolor, label={[font=\scriptsize\sffamily, xshift=0.15em]right:Kip. Ring}, inner sep=0.2em] at       (0,1){};
\node[circle, fill=blackforestcolor, label={[font=\scriptsize\sffamily, xshift=0.15em]right:Black For.}, inner sep=0.2em] at   (0,2){};
\node[circle, fill=mumbaicolor, label={[font=\scriptsize\sffamily, xshift=0.15em]right:Mumbai}, inner sep=0.2em] at              (1,0){};
\node[circle, fill=capetowncolor, label={[font=\scriptsize\sffamily, xshift=0.15em]right:C. Town}, inner sep=0.2em] at         (1,1){};
\node[circle, fill=chabarovskcolor, label={[font=\scriptsize\sffamily, xshift=0.15em]right:Chabar.}, inner sep=0.2em] at      (1,2){};
\node[circle, fill=mexicocitycolor, label={[font=\scriptsize\sffamily, xshift=0.15em]right:Mex. C.}, inner sep=0.2em] at     (1,3){};
\end{tikzpicture}
        }
    
        \begin{tikzpicture}[inner sep=0]
            \tikzstyle{legendentry}=[]
            \node[](img) {\includegraphics[width=\textwidth]{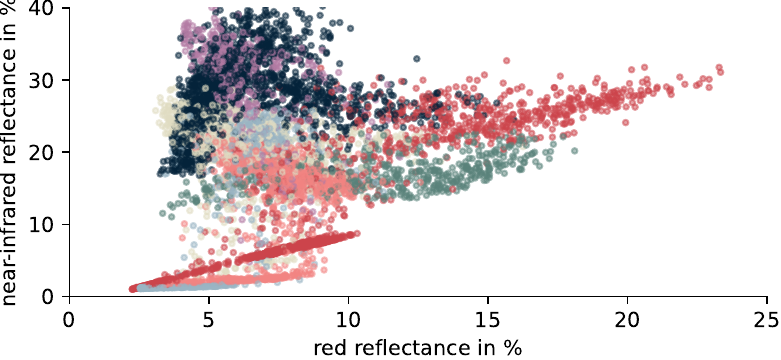}};

            \node[fill=none, fill opacity=0.5, text opacity=1, rounded corners] at (1.75,-.45){\legend};

        \end{tikzpicture}
        \caption{Geographic domain shift in red-nir feature space. Points represent the mean reflectance in each image}
    \label{fig:exp1:scatter}
    \end{subfigure}
    \hfill
    \begin{subfigure}{.49\textwidth}
                \newcommand{\domainlegend}{%
\begin{tikzpicture}[xscale=1.8, yscale=0.25]
\node[circle, fill=sourcedomaincolor, label={[font=\scriptsize\sffamily, xshift=0.15em]right:source images}, inner sep=0.2em] at (0,1){};
\node[star, star points=5, star point height=0.2em, rounded corners=0, fill=targetdomaincolor, label={[font=\scriptsize\sffamily, xshift=0.15em]right:target images}, inner sep=0.1em] at (0,0){};

\node[circle, draw=sourcedomaincolor, label={[font=\scriptsize\sffamily, xshift=0.15em]right:sample weight}, inner sep=0.2em] at (0,2){};
\node[circle, draw=sourcedomaincolor, inner sep=0.1em, yshift=-.1em] at (0,2){};
\node[circle, draw=sourcedomaincolor, inner sep=0.3em, yshift=.2em] at (0,2){};

\end{tikzpicture}
        }
        \begin{tikzpicture}[inner sep=0]
            \node[]{\includegraphics[width=\textwidth]{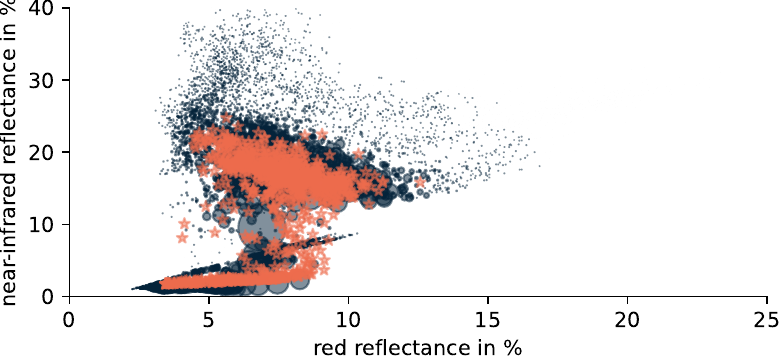}};

            \node[fill=none, fill opacity=0.5, text opacity=1, rounded corners] at (2.2,-.45){\domainlegend};
            
        \end{tikzpicture}
        
        \caption{Source images weighted by relevance to match the Mumbai target region}
    \label{fig:exp1:scatter_weighted}
    \end{subfigure}
    
    \begin{subfigure}{.6\textwidth}
        \centering\includegraphics[width=\textwidth]{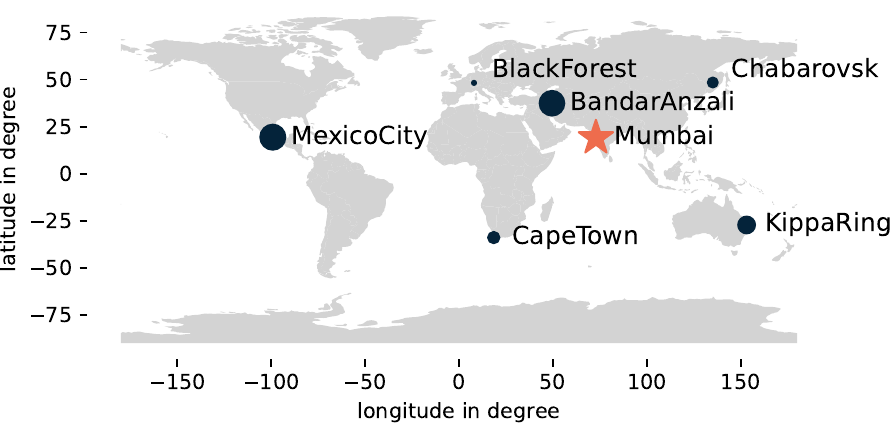}
        \caption{Target area Mumbai, samples from Mexico City and Bandar Anzali receive the highest importance score indicated by size}
    \label{fig:exp1:map}
    \end{subfigure}
    \hfill
    \begin{subfigure}{.35\textwidth}
        \centering\includegraphics[width=\textwidth]{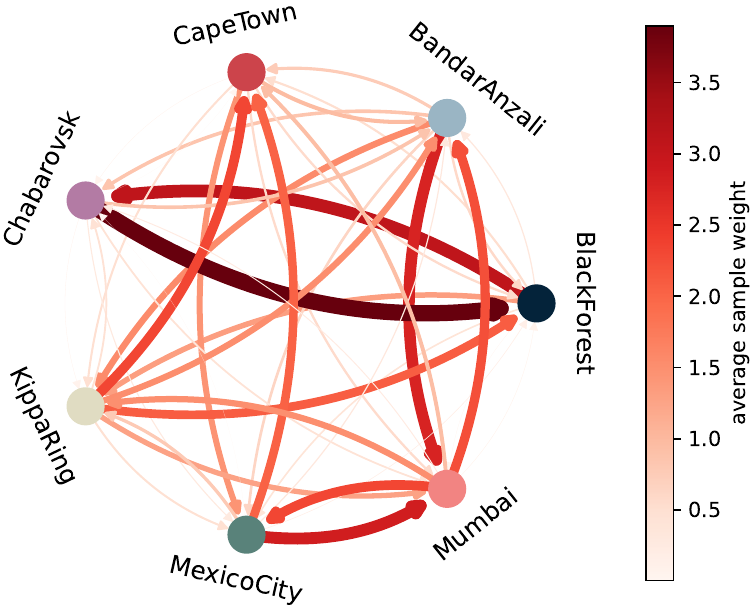}
        \caption{Cross-region sample average importance score between DFC regions}
    \label{fig:exp1:graph}
    \end{subfigure}

        
    
    \caption{Study 1. Analyzing geographic domain shift with KLIEP to derive a relevance-weighting scheme for model training.}
    \label{fig:exp1}
\end{figure*}

\noindent
\textbf{Motivation}. 
Varying environmental and geographic conditions changes the visual representation of land cover classes. This can be seen in \cref{fig:exp1:scatter} where the average reflectance in red and near-infrared channels varies systematically between the DFC2020 regions indicated by color.
This manifests a data distribution shift between geographic regions, negatively impacting model classification accuracy when not all training data is equally \emph{relevant} for the particular testing region. In this case, data from the source distribution (i.e., training data) does not overlap with the target distribution (testing data). This can be seen in the case of Cape Town and Black Forest in \cref{fig:exp1:scatter}.

\noindent
\textbf{Aim}.
We aim to reduce the distribution shift with respect to a target region by weighting each image sample with an importance coefficient in an instance-based transfer learning approach \cite{pan2010survey}. In this method, samples from the source distribution that are assigned high importance weights are deemed more \emph{relevant} for the particular target distribution, while low weights are assigned to irrelevant samples.

\noindent
\textbf{Method}.
We use the Kullback–Leibler importance estimation procedure (KLIEP) \cite{sugiyama2007direct} using the Frank-Wolfe algorithm \cite{wen2015correcting} implemented in the Python Adapt package \cite{de2021adapt} to determine the relevance-weighting scheme. 
This algorithm approximates the source and target distributions by the sum of each sample's radial basis function (RBF) kernel densities. We perform a grid-search for kernel width between 0.1 and 2.0 in regular 0.1 steps. KLIEP iteratively adjusts the sample weights so that Kullback-Leibler (KL) divergence between target distribution and source distribution is minimized, as shown in \cref{fig:exp1:scatter_weighted}.
We represent each image sample as a 20-dimensional vector of pixel-average band reflectances (10 channels) and their standard deviations (10 channels).
We then fit a random forest classifier from Scikit-Learn \cite{scikit-learn} with uniform weighting and a KLIEP-derived importance weighting on the source regions and measure the accuracy on the target region. 


\noindent
\textbf{Results and Interpretation}. 
We show results in \cref{fig:exp1:map,fig:exp1:graph,fig:exp1:table}. 
First, \cref{fig:exp1:map} shows a world map with average relevance of all images within each source DFC region (scaled circles) with respect to Mumbai. This generally covers Tobler's \cite{Tobler1970} intuition that more geographically nearby areas, like Bandar Anzali, are more relevant than distant ones. However, environmental conditions beyond mere geographic distance also contribute to the relevance, as the high importance of Mexico City images shows that it shares a similar arid climate.
Second, \cref{fig:exp1:graph} extends this comparison across all regions and shows the sample-averaged importance scores as edge width and color in a bi-directional graph between all DFC different regions. Here, we can see more generally that nearby (Mumbai $\leftrightarrow$ Bandar Anzali) temperate and forested regions (Black Forest $\leftrightarrow$ Chabaraovsk) and arid regions (Cape Town $\leftrightarrow$ Mexico City $\leftrightarrow$ Mumbai $\leftrightarrow$ Kippa Ring) are more relevant for each other.

\begin{table*}
 \centering
        \caption{Target accuracy in [\%] using regular-uniform weighting vs. relevance-weighting of training data}
        \label{fig:exp1:table}
        \begin{tabular}{lrrrrrrr}
            \toprule
Region & B. Anz. & Mumbai & Mex. C. & C. Town & B. For. & Chab. & Kip. R. \\
            \cmidrule(lr){2-2}\cmidrule(lr){3-3}\cmidrule(lr){4-4}\cmidrule(lr){5-5}\cmidrule(lr){6-6}\cmidrule(lr){7-7}\cmidrule(lr){8-8}
            Uniform sampling & 54.73 & 49.78 & 60.33 & 81.11 & 69.04 & 25.93 & 58.63 \\
            Weighted sampling & 55.47 & 50.78 & 59.92 & 80.90 & 75.35 & 28.19 & 59.80 \\
            \cmidrule(lr){2-8}
            Difference & 0.74 & 1.00 & -0.41 & -0.21 & 6.30 & 2.26 & 1.17 \\
            \bottomrule
        \end{tabular}
\end{table*}
    
Finally, Table \ref{fig:exp1:table} quantifies the accuracy benefit of classifying on a sample-weighted training source dataset with respect to each target region in the columns. The row \enquote{uniform weighting} shows classification results on the random forest trained on regular uniformly weighted training data, while \enquote{relevance weighting} reports the accuracy on KLIEP-weighted training samples. The \enquote{difference} rows indicate in which regions the de-biasing was beneficial (positive) or detrimental (negative) to the final target accuracy.
Here, the weighting positively affected five of the seven DFC regions up to 6.3\%. Two regions were classified at lower accuracy (up to -0.41\%). That this procedure can also have detrimental effects in same case illustrates the inherent challenge of deriving the relevance of samples and that negative transfer can occur and deteriorate performance.

\noindent
\textbf{Summary}.
We implement and demonstrate the KLIEP algorithm to derive a relevance-based weighting scheme to account for geographic distribution shifts. This reveals patterns in the data, such as which geographic areas are more \emph{classification-relevant} than others, but also improves the predictive accuracy when weighting the training samples accordingly. However, this procedure does not always increase the accuracy. It can also have detrimental effects when less relevant samples are falsely assigned a higher importance/relevance score due to incomplete approximations and simplifications, as is necessary for choosing kernel functions or the distribution divergence metric with computational restrictions in mind. Further research is necessary to make these algorithms applicable with less supervision and more general with less hyperparameter tuning.

\subsection{Study 2: Pruning noisy data with confident learning}
\label{sec:exp2}

\begin{figure*}
     \centering
     \begin{subfigure}[b]{0.5\textwidth}
         \centering
         \includegraphics[width=\textwidth]{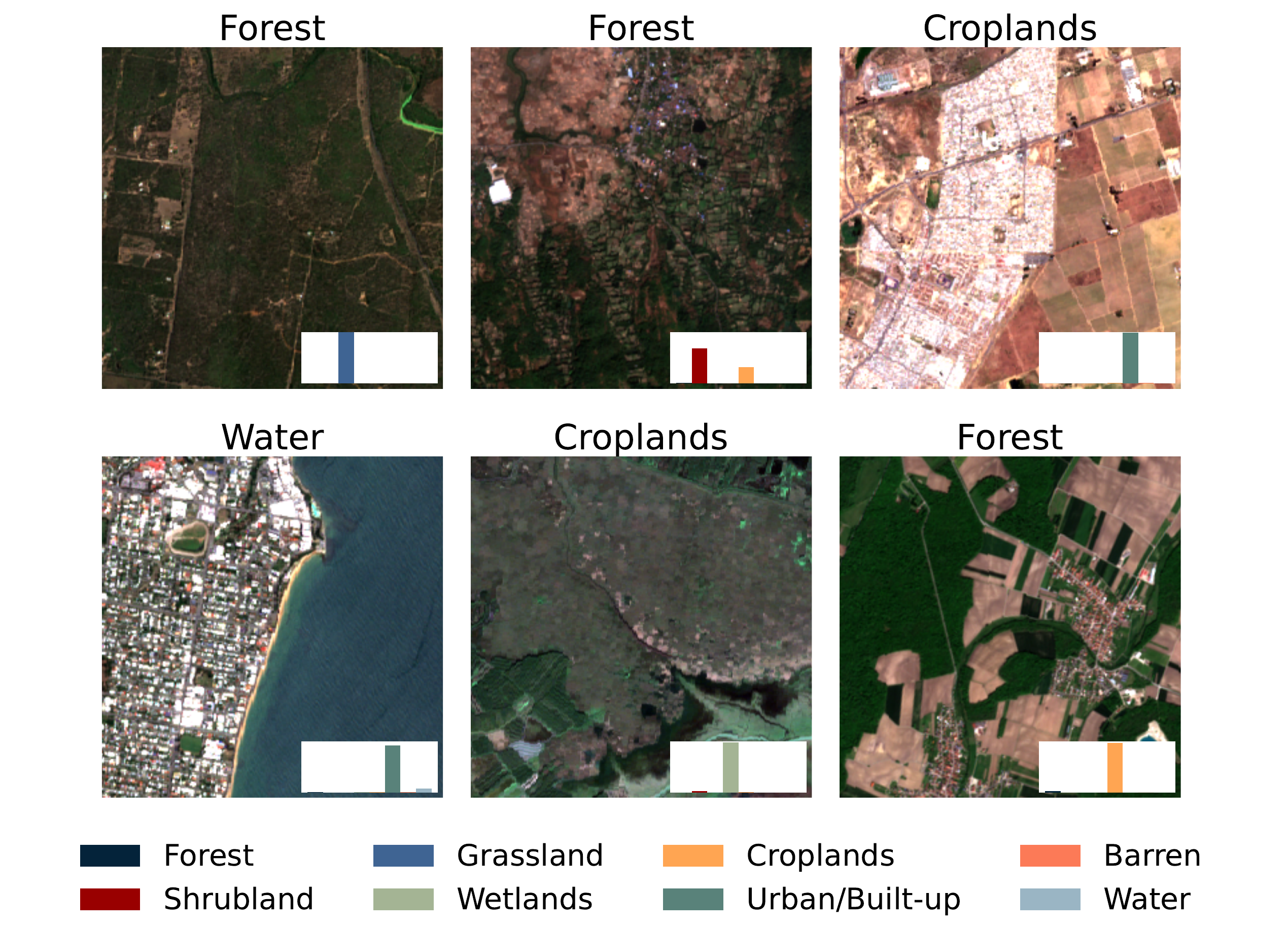}
         \caption{}
         \label{fig:exp2:noisy_labels}
     \end{subfigure}
     \hfill
     \begin{subfigure}[b]{0.49\textwidth}
         \centering
         \includegraphics[width=.9\textwidth]{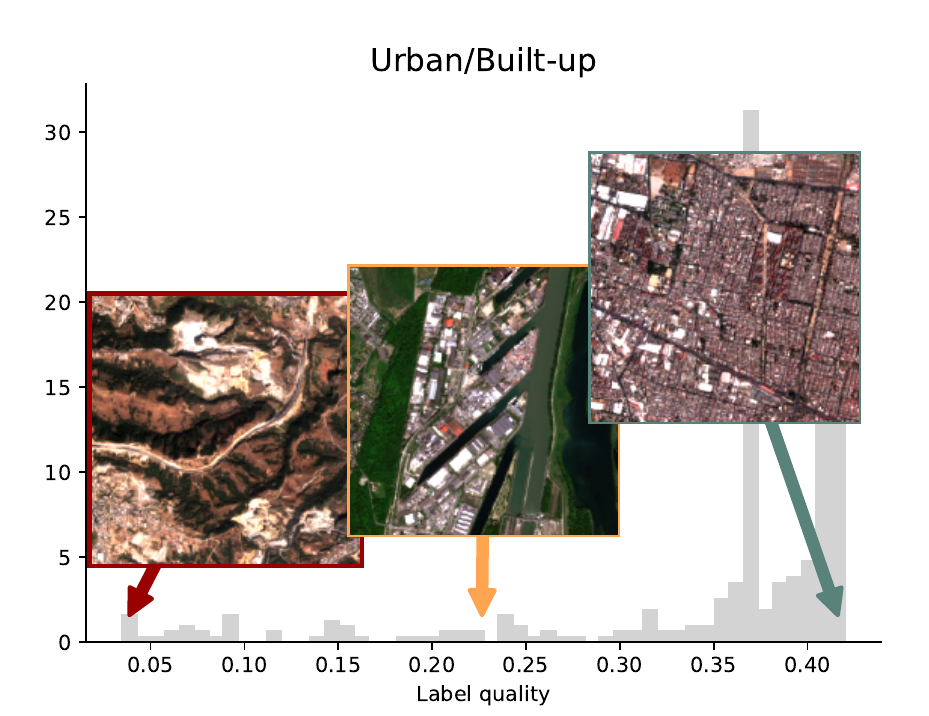}
         \caption{}
         \label{fig:exp2:label_quality}
     \end{subfigure}
     \caption{Qualitative results, Experiment 2. (a) Potential label issues detected by Confident Learning in the DFC training set. The model's predicted class probabilities are shown as bar charts. (b) Histogram of the label quality of class \enquote{Urban/Built-up} with three example images.}
\end{figure*}

\begin{table}[ht]
    \centering
    \caption{Quantitative results of study 2. Model performance when training on the entire training dataset (baseline) and after pruning noisy training samples (confident learning).}
    \label{tab:exp2}
    \begin{tabular}{cc}
    \toprule
          & Accuracy [\%]\\
          \midrule
         Baseline & 66.53\\
         Confident learning & 69.98\\
    \bottomrule
    \end{tabular}
\end{table}

\noindent
\textbf{Motivation}.
\emph{Accurate} data labels are crucial for the effectiveness of trained models, yet inaccuracies can significantly undermine model performance. In geospatial remote sensing, label issues often arise due to similarities in appearance between classes, such as \enquote{Grassland} and \enquote{Shrubland} sharing comparable vegetation characteristics, which can lead to confusion. Additionally, the spatial distribution of certain classes, like \enquote{Wetlands} being commonly located near water, increases the likelihood of co-occurrence within patches, further complicating accurate labeling. 

\noindent
\textbf{Aim}.
Our aim is to improve label accuracy by ranking data samples and pruning potentially noisy samples

\noindent
\textbf{Method}.
To achieve this, we employ confident learning~\cite{northcutt2019confidentlearning}, a framework designed to estimate label uncertainty and enhance dataset quality. Confident learning operates under the assumption that label noise is class-conditional, depending solely on the true class rather than the data itself~\cite{angluin1988learning}. Although this is a simplification, it is a reasonable assumption for many scenarios, including land cover classification.

The method involves training a machine learning model on the dataset and then estimating the joint distribution of noisy (given) labels and true (unknown) labels. This estimation is done by identifying examples likely to have label issues. For instance, it counts all patches labeled as \enquote{Grassland} that have a sufficiently high predicted probability of being \enquote{Shrubland}. Here, a value for sufficiently high is determined by a class-specific threshold calculated as the model's average self-confidence for each class. Using the estimated joint distribution, all samples are ranked based on their likelihood of being noisy. Potentially noisy samples are then pruned from the dataset.

\noindent
\textbf{Results and Interpretation}. 
The confident learning method flagged 123 samples in the DFC training set as having potential label issues. Six of these samples are presented in~\cref{fig:exp2:noisy_labels}, illustrating instances where the model disagreed with the given label. In some cases, as seen in the lower left image, discrepancies arise from the presence of multiple classes within one patch, while in other cases, such as the top left image, the reason is related to the similarity in appearance between classes.
\Cref{fig:exp2:label_quality} shows a histogram of the label quality of all training samples belonging to class \enquote{Urban/Built-up}, accompanied by three example images representing different label qualities: one with high label quality, one with medium label quality, and one with low label-quality. These examples show that the label quality tends to degrade as the fraction of \enquote{Urban/Built-up} pixels within the patch decreases.
To improve the quality of the training data, the samples flagged as noisy were removed. Re-training the model on this pruned dataset resulted in an overall accuracy improvement of +3.45 percentage points compared to the baseline (see Table~\ref{tab:exp2}).

\noindent
\textbf{Summary}.
Confident learning is a form of data curation that is typically used to improve the training dataset quality by pruning noisy data. This approach is especially relevant for settings with low label granularity, where multiple classes are present within the same image. By pruning the dataset, confident learning aims to improve the \emph{accuracy} of the labels. 


\subsection{Study 3: Evaluation with slice discovery}
\label{sec:SliceDiscoveryMethodes}

\noindent
\textbf{Motivation}.
To ensure a comprehensive evaluation of model performance, it is beneficial to identify human-understandable subsets of test samples within each class, where the model's performance varies across different slices. This approach provides deeper insights into the model's strengths and weaknesses.

\noindent
\textbf{Aim}.
Our aim is to demonstrate the insights that can be gained by analyzing slices in the test data.

\noindent
\textbf{Method}.
We utilize the idea of slice-finding methods, such as presented in \cite{eyuboglu2022domino}. 
To analyze the data at test time, we first represent all data points by a latent representation vector, produced via feature extraction of a pre-trained \textit{ResNet18} neural network of Prexl et al. \cite{prexl2023multi}, which is a domain-specific (trained on \textit{Sentinel-2} images of the \textit{SEN12MS} dataset) pre-trained network based on the SimCLR contrastive learning paradigm \cite{chen2020simclr}. 
We extract a $512$ dimensional feature for each image in the test set of the \textit{DFC2020} dataset.
Using those latent feature representations allows for sub-clustering of all samples within one class by applying \textit{k-means} clustering in the latent space.

\begin{figure*} 
     \centering
     \begin{subfigure}[b]{0.4\textwidth}
         \centering
         \includegraphics[width=\textwidth]{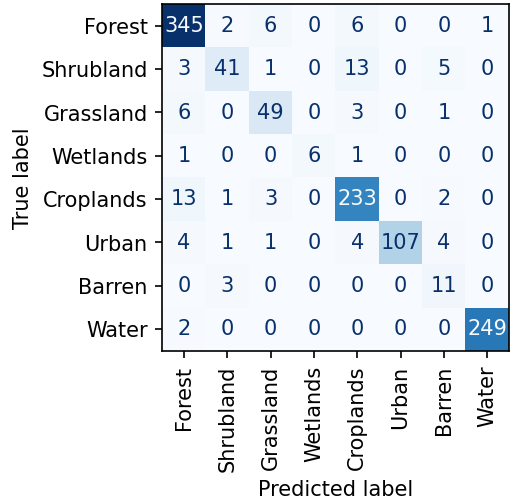}
     \end{subfigure}
     \begin{subfigure}[b]{0.4\textwidth}
         \centering
         \includegraphics[width=\textwidth]{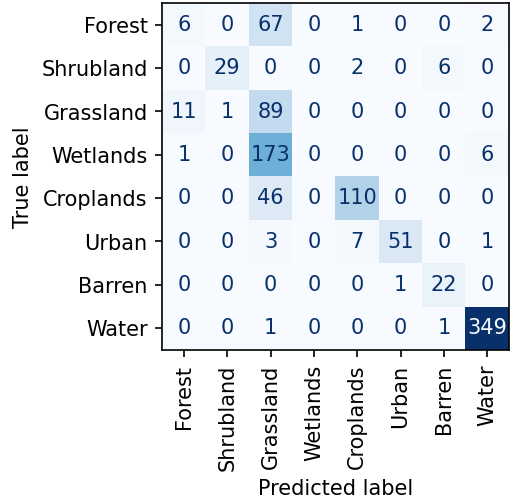}
     \end{subfigure}
    \caption{The confusion matrices (left: validation set, right: test set) of the model described in \cref{sec:expsetup} for validation (left) and test set (right).}
    \label{fig:exp4_cm}
\end{figure*}

\begin{figure*} 
     \centering
     \includegraphics[width=0.7\textwidth]{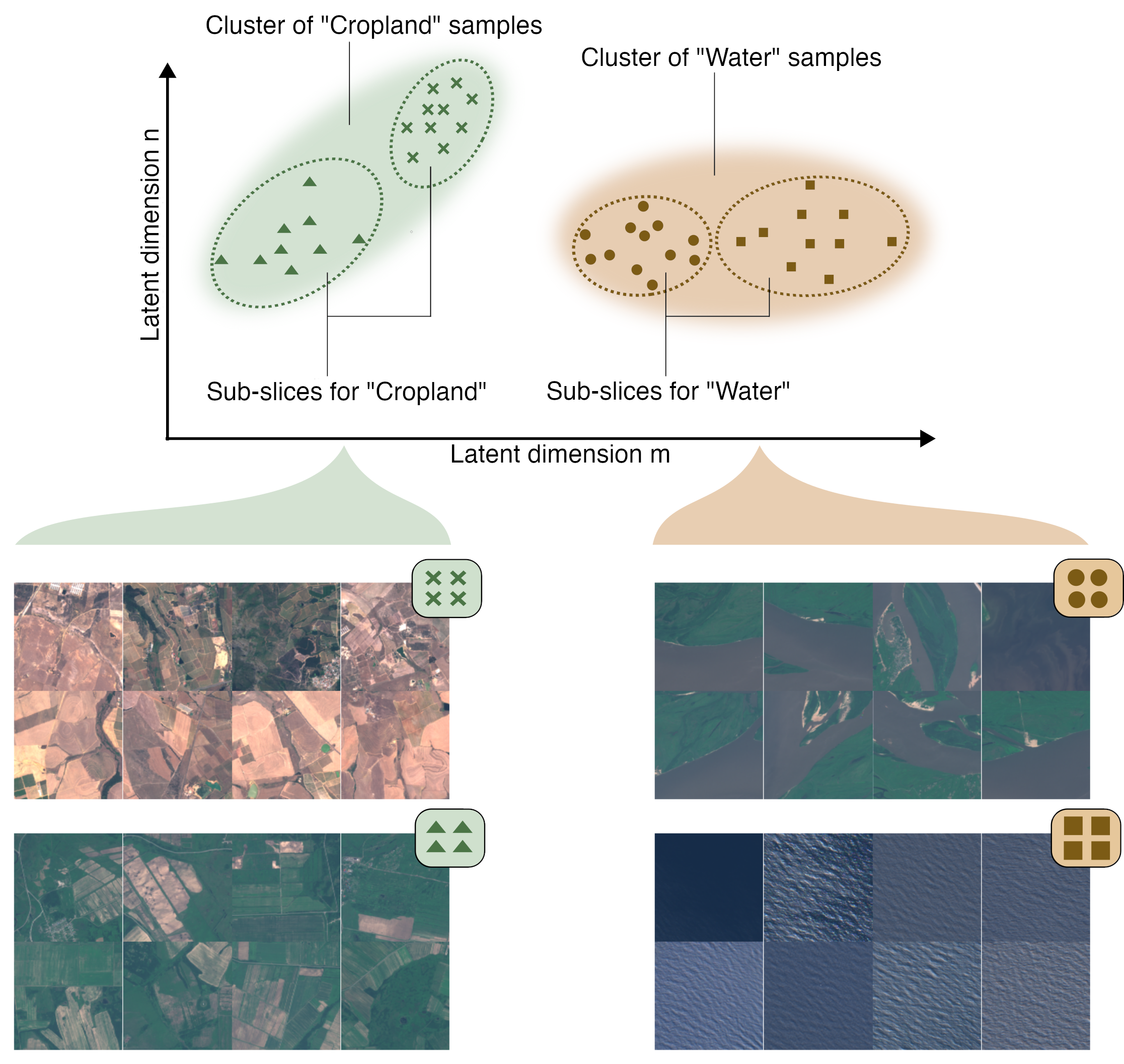}
     \caption{The cluster analysis showcases representative slices of all samples for the "Cropland" and "Water" classes. By clustering the latent representations, distinct sub-clusters of cropland and water types are identified within the test set of the dataset. The top row presents a schematic example of the clustering results, while the bottom two rows offer visual examples of the sub-cluster analysis for the \enquote{Cropland} and \enquote{Water} classes.}
     \label{fig:exp4}
\end{figure*}

\noindent
\textbf{Results and Interpretation}. 
The cluster-center affiliation of each sample allows for quick and robust visual analysis of the data and can produce further insights about the task at hand. This can subsequently be used to make informed decisions regarding improved labeling schemes or feedback into the data curation step (Step 2 in \cref{fig:pipeline}).
We show two examples of \enquote{Cropland} and \enquote{Water} with their dominant sub-slices in \cref{fig:exp4}.
This reveals directly different patterns within one class label, which might harm the model's performance.
In the first case, looking at all samples labeled as \enquote{Water}, it can be observed that the slices capture different subcategories like river, lake, and open ocean.
While \enquote{Water} is accurately classified, \enquote{Cropland} is commonly confused with \enquote{Grassland}, as shown in the confusion matrix \cref{fig:exp4_cm}. Here, analyzing sub-slices can reveal failure cases: \enquote{Cropland} samples cluster in two slices with different climate conditions or seasons. While the performance on all samples in the first slice (\cref{fig:exp4}) has $100\%$ accuracy, the performance for the second slice drops significantly to $10\%$, likely due to the high similarity to the \enquote{Grassland} samples it he training set.

\noindent
\textbf{Summary}.
Overall, slice discovery methods serve as tools to gain insights into the outcome of a specific task that is given to the model beyond accumulated numbers. This insight can be used directly for data creation and curation to improve the performance by, for example, creating new classes for the sub-slices. 

\section{Discussion \& Conclusion}
Shifting research focus from well-explored model-centric approaches towards the broader aspects of the entire machine learning cycle, especially data-centric learning techniques, is necessary to make further progress towards accurate and generalizable models. Here, we find that so far, only limited research is focused on automated data-centric approaches that explicitly improve the quality criteria that we categorized in \cref{sec:terminology} or utilize them for the model-building process. In this work, we provided an extensive literature review in \cref{sec:steps} and categorized the techniques according to steps in the machine learning cycle and identified the quality criteria they improve or use. This literature study was complemented by concrete implementation examples in \cref{sec:experiments} to stress the applicability of some of the data-centric methods. 

\vspace{1em}
\noindent
\textbf{Research Gaps}. Unlike the numerous well-established metrics available to measure the performance of machine learning models, standardized metrics to evaluate the quality of a dataset are not established so far \cite{aroyo2022data}. Within the studied data-centric approaches, we identified research gaps in \cref{fig:techniques_dotplot} in the following steps: 
\begin{description}
    \item[Step 2: Creation] Only a few existing works focus on automated data creation to improve data consistency, unbiasedness, or relevance. While these criteria are often considered in the manual selection of training and evaluation sites, little work has been found on the automation of this process through data-centric learning in the geospatial domain. Also, little research has been conducted so far on the benefit of synthetic, generated, and augmented data for geospatial remote sensing data.  
    \item[Step 3: Curation] We find that few works focus on maximizing diversity, consistency and/or unbiasedness through data curation. This has several factors: First, it is plausible for diversity, as curation is more aimed at cleaning up and deleting data, which hardly increases diversity. Moreover, it is challenging to quantify diversity and completeness as a reference is missing. Usually, such shortcomings are revealed after the model deployment step, which can be fed back to the machine learning cycle. Currently, however, most works do not consider feedback loops. 
    Research related to consistency is restricted due to the limited availability of datasets (i.e., multi-labeler datasets) that allow focusing on this aspect. Bias is still an under-investigated criterium and its cascading effects on the results are underestimated. Usually, biases in the dataset are revealed too late in the model deployment (Step~6). Fortunately, a new sensitivity has evolved towards this criteria. Overall, research is currently hindered due to the lack of clear objectives on what a data set should look like regarding the quality criteria. For example, curation strategies can be too time consuming and costly compared to their benefit on the dataset. Also data-centric and model-centric developments need to be balanced to find the best way to deal with the data and the application.   
    \item[Step 4: Utilization] We find that few works use completeness or unbiasedness. However, in contrast to previous steps, we believe that these criteria are less relevant for the data utilization in the model training context, while diversity, accuracy, and relevance are well covered by existing literature. 
    \item[Step 5: Evaluation] As already underlined by Rolf et al. \cite{rolf2023evaluation}, there are several opportunities for a better evaluation that have not been exploited yet. Investing in a high-quality evaluation data set is important to get a comprehensive view on the ability of the model. Further, we need to distinguish between performance measures reported with a statistical parameter and measures that provide insights into the skill for a given task. Generally, there is limited research that uses consistency, bias, and relevance information of the data to get insights into the model's performance. 
\end{description}

\vspace{1em}
\noindent
\textbf{Experimental Performance}. 
Another important point of discussion is the effectiveness of current data-centric approaches in experiments. While all proposed methods are well-motivated, the actual improvement in accuracy often falls short of expectations. For example, in Study 1 (Section \ref{sec:exp1}), weighting dataset samples regarding their relevance only improves two out of five regions. Similarly, in Study 2 (Section \ref{sec:exp2}), the quantitative accuracy gain by removing samples with label noise only results in a 3.5\% improvement. However, both approaches reveal underlying patterns in the data that can provide valuable insights on their own. Additionally, in real-world applications, accuracy is not the sole factor of interest. Factors such as reliability, precision, robustness, and adaptability are often equally important. Furthermore, a major goal is to avoid the negative effects caused by low-quality data \cite{sambasivan2021everyone}. We further want to point out that the used methods were chosen due to their already proven performance in diverse fields. However, the success of data-centric techniques also depend on the chosen model, therefore we encourage considering the whole machine learning cycle, including model- and data-centric aspects.    

\vspace{1em}
\noindent
\textbf{Limitations}. 
Several limitations need to be considered in the interpretation of this study. First, the proposed categorizations are not sharp. Methods can be argued for improving multiple criteria. For instance, the first study uses KLIEP \cite{sugiyama2007direct} (data curation) to address the relevance, but it can also be seen as a de-biasing strategy. Here, the irrelevant samples with respect to one target area receive a low weight and are discarded. Simultaneously, this process of selecting the most relevant instances de-biases the entire datasets if we are aware of a certain sampling bias, e.g., towards some continents. Similarly, the process of data augmentation can be argued towards data creation (step 2), as new augmented samples are virtually created, or data utilization (step 4), as the existing data is utilized more effectively by flipping and rotating images without the need to collect new data samples.
Finally, we highlight that the manual search of literature always introduces a certain selection bias. We aimed to mitigate this by inviting a diverse range of co-authors who work across disciplines that create, curate, and utilize geospatial remote sensing data.
We believe that this broad expertise allowed us to extensively and intensively search the entire body of published literature in this domain.

\vspace{1em}
\noindent
\textbf{Conclusion and Outlook}. 
We consider this review as a starting point to raise awareness about the possibilities of data-centric machine learning in the geospatial domain rather then arguing against model-centric research. This review introduced the definitions, and relevant criteria to give researchers the tools and structure to improve datasets in an automated and systematic way. Focusing on the entire machine learning cycle from problem definition to model deployment with feedback is necessary to train machine learning models that can generalize well to new locations and can be trusted in unexpected situations. This is captured and quantified in data-centric learning that explicitly improves and uses the relevant data quality criteria. Future research is necessary to fill the identified research gaps in data creation and curation towards a higher quality of datasets. Further, additional methodological developments are necessary to improve the reliability of data-centric methods that today often fall behind expectations. In summary, we hope that this review helps to structure and guide this discussion and leads to method developments targeted towards new unexplored directions that help tackle real-world problems with high-quality data.

\section{Acknowledgments}
The contribution of Ribana Roscher is partially funded by the DFG under Germany’s Excellence Strategy – EXC 2070 – 390732324.
The contribution of Caroline Gevaert is partially funded through the project “Bridging the gap between Artificial Intelligence and society: Developing responsible and viable solutions for geospatial data” (with project number 18091) of the research program Talent Programme Veni 2020 by the Dutch Research Council (NWO). The contribution of Michael Kampffmeyer and Stine Hansen is partially funded by the Research Council of Norway through its Centre for Research-based Innovation funding scheme (grant no. 309439). The contribution of Jefersson A. dos Santos and Keiller Nogueira is partially funded by the Serrapilheira Institute (grant R-2011-37776).

\bibliographystyle{ieeetr}
\bibliography{references}

\end{document}